\documentclass[letterpaper]{article} 
\usepackage{aaaiWorkshop}  
\usepackage{times}  
\usepackage{helvet}  
\usepackage{courier}  
\usepackage[hyphens]{url}  
\usepackage{graphicx} 
\usepackage{natbib}  
\usepackage{caption} 
\usepackage{algorithm}
\usepackage{algorithmic}

\usepackage{newfloat}
\usepackage{listings}
\DeclareCaptionStyle{ruled}{labelfont=normalfont,labelsep=colon,strut=off} 
\floatstyle{ruled}
\newfloat{listing}{tb}{lst}{}
\floatname{listing}{Listing}
\title{An Agentic Approach to Automatic Creation of P\&ID Diagrams from Natural Language Descriptions}
\author {
    Shreeyash Gowaikar\textsuperscript{\rm 1},
    Srinivasan Iyengar\textsuperscript{\rm 2},
    Sameer Segal\textsuperscript{\rm 1},
    Shivkumar Kalyanaraman\textsuperscript{\rm 2}
}
\affiliations {
    \textsuperscript{\rm 1}Microsoft Research India\\
    \textsuperscript{\rm 2}Mircosoft Corporation\\
}

\usepackage{bibentry}

\begin{document}

\maketitle

\begin{abstract}
The Piping and Instrumentation Diagrams (P\&IDs) are foundational to the design, construction, and operation of workflows in the engineering and process industries. 
However, their manual creation is often labor-intensive, error-prone, and lacks robust mechanisms for error detection and correction. 
While recent advancements in Generative AI, particularly Large Language Models (LLMs) and Vision-Language Models (VLMs), have demonstrated significant potential across various domains, their application in automating generation of engineering workflows remains underexplored.
In this work, we introduce a novel copilot for automating the generation of P\&IDs from natural language descriptions.
Leveraging a multi-step agentic workflow, our copilot provides a structured and iterative approach to diagram creation directly from Natural Language prompts.
We demonstrate the feasibility of the generation process by evaluating the soundness and completeness of the workflow, and show improved results compared to vanilla zero-shot and few-shot generation approaches.
\end{abstract}

%

\section{Introduction}
\label{intro}

\begin{figure*}[ht]
    \centering
    \includegraphics[width=0.75\linewidth]{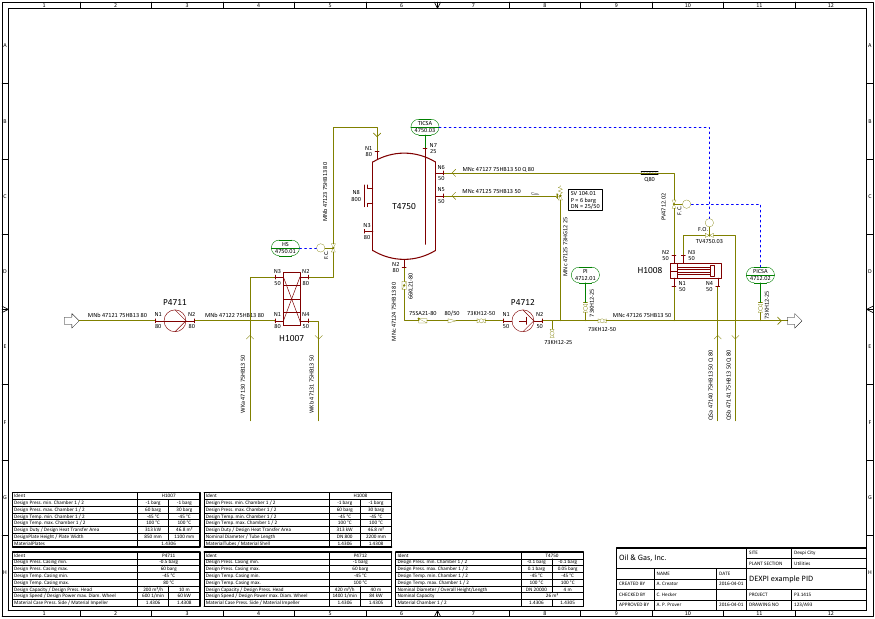}
    \caption{An example of a simple P\&ID sheet \cite{dexpi-pid-example}. It entails Equipments (Tank, Heat Exchangers and pumps), Piping Systems, Instruments (Globe Valves, Ball Valves, Pipe Reducers, Butterfly Valve, Spring Loaded Globe Safety Valve and Swing Check Valves), Pipe Off Page Connectors and Actuating Systems. 
    }
    \label{fig:PID_Example}
\end{figure*}

Engineering Diagrams (EDs) such as Block Flow Diagrams (BFDs), Process Flow Diagrams (PFDs) and Piping and Instrumentation Diagrams (P\&IDs) are essential for the design, construction, operation, and maintenance of chemical and process plants \cite{sakhina-vqa-llms}.
They play a crucial role in the effective dissemination of information to all stakeholders in the chemical, energy and process industry.
BFDs show the preliminary design of unit operations, process flows, inputs, and outputs, while PFDs present major equipment, flow paths, key instruments, and some process details like flow rates and contents, and P\&IDs offer a more detailed view, showing all equipment, pipes, and instruments in a plant \cite{nasby-flowsheet}.
However, the process of creating PFDs and P\&IDs manually, relying on schemes from previous projects, design guidelines, engineer expertise, and Computer-Aided Design (CAD) software, can often become tedious and error-prone \cite{Hirtreiter-psd2pids}.
In addition to being laborious, due to the manual creation, the entire diagram generation process has minimal provenance for error detection and correction. 

In the past few years, Generative AI has achieved a remarkable progress.
Large Language Models (LLMs) now excel in tasks like Text Classification, Question Answering, Text Summarising, Text Translation, Machine Translation, Code Generation, Text Generation, and Sentiment Analysis \cite{laskar-etal-2023-systematic-eval}.
Multi modal Vision Language Models (VLMs), have also progressed to solve tasks like Visual Reasoning, Visual Question Answering and Image Generation \cite{zhang2024visionlanguagemodelsvisiontasks}.
Attributed to this progress, Generative AI has been adopted for various applications and automations across a myriad of industries and domains \cite{gozalobrizuela2023surveygenerativeaiapplications}.

Artificial Intelligence (AI) has been employed in the chemical and process industry for over 35 years and it has achieved notable successes \cite{venkat-aiche-app}.
The earliest use of AI in process systems engineering, include the AIDES (Adaptive Initial DEsign Synthesizer system) \cite{siirola-aides}, that uses end-to-end analysis, linear programming and symbol matching for chemical process generation, the Design-Kit \cite{STEPHANOPOULOS-designkit}, a rule-based logical program for flow sheet synthesis and operational analysis, MODELL.LA \cite{STEPHANOPOULOS-modellla}, a modelling language for process system models, and MODEX \cite{RICH-modex}, a causal model based system for fault diagnosis. 

More recently, Machine Learning (ML), Deep Learning (DL) and Reinforcement Learning (RL) algorithms are being used to solve classical problems in the process industry. 
There are efforts to use ML, DL, and RL algorithms to digitize classical P\&IDs in paper or PDF format \cite{paliwal_autoPID, chul_e2eDigitization}, predict next element in PFDs and P\&IDs \cite{vogel_sfiles_to_flowsheets}, translate PFDs to P\&IDs \cite{autoPID_pfds}, generate control code \cite{koziolek_control}, and even do question answering on P\&IDs \cite{sakhina-vqa-llms}.
However, we find a gap in existing literature to automate  generation of P\&IDs directly from Natural Language.
Our work serves as a preliminary to the field of generating P\&IDs directly from Natural Language utterances.

Through this work, we introduce a novel application of Generative AI for the automatic, sub-system level generation of P\&IDs.
We propose a copilot powered by a multi-step agentic workflow to generate a P\&ID from only its linguistic description.
We nickname the copilot - the ACPID Copilot as an acronym for the \textbf{A}utomatic \textbf{C}reation of \textbf{P}\&\textbf{ID}s.
Aimed at improving efficiency and productivity of the engineers, the ACPID copilot also improves provenance of the generation process by enabling audit trails.  
The copilot outputs a textual representation of the P\&ID, a Microsoft Visio Diagram of the P\&ID for engineers to edit during the generation process, and also a natural language description of the current P\&ID that can be append in subsequent prompts for iterative development through multi-turn conversations. 
The copilot's design also allows users to edit and start with an existing P\&ID, facilitating quick adoption and integration into existing workflows.
We have organised the paper into the following sections for easier understanding:
\begin{itemize}
    \item The \emph{Background} section provides the essential background information.
    \item The \emph{Methodolgy} section describes in detail the entire process design. 
    \item The \emph{Evaluations} section describes the evaluation strategy and provides the results of our approach.
    \item We conclude with the \emph{Discussions} and \emph{Conclusions} section, which talks about the impact, limitations and future avenues of our research.
\end{itemize}

\begin{figure*}[ht]
    \centering
    \includegraphics[width=\linewidth]{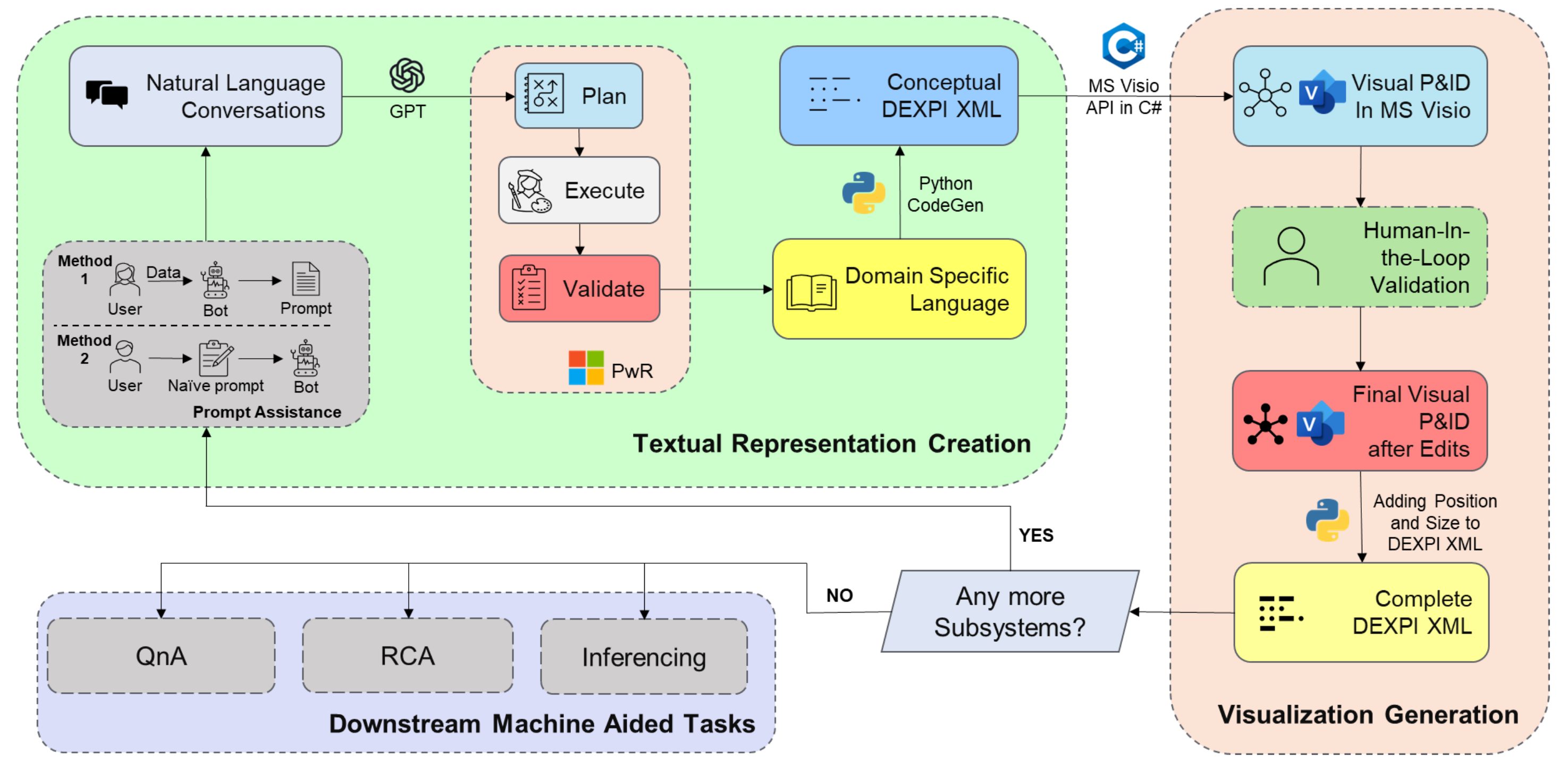}
    \caption{The architectural block diagram of the ACPID Copilot 
    }
    \label{fig:arch}
\end{figure*}

\section{Background}

\subsection{P\&ID}
\label{bg:pid}

Piping and Instrumentation Diagrams (P\&IDs) are critical tools for the design, operation, and maintenance of complex process systems, including those found in chemical plants, refineries, and power stations. 
An example of a P\&ID diagram can be seen in Figure \ref{fig:PID_Example}. 
These diagrams serve as blueprints, providing a comprehensive view of the processes, equipment, and instrumentation
They are integral to various aspects of system management, including financial planning, safety considerations, and operational efficiency.
\paragraph{Financial Impact} P\&IDs directly influence financial decisions by enabling the creation of accurate Bills of Material (BoM) and optimizing supply chain planning.  
\paragraph{Safety Considerations} P\&IDs play a crucial role in ensuring safety, particularly in environments where hazardous materials or extreme conditions are involved. 
By detailing the flow of materials and equipment interconnections, P\&IDs help in developing safety protocols and hazard mitigation strategies.  
\paragraph{Operational Planning} P\&IDs are essential for managing ongoing operations and maintenance by ensuring all system components are well-documented for efficient decision-making. 
They also aid in planning project timelines and inventories, and are particularly valuable for retrofitting or upgrading facilities, providing a clear reference for modifications with minimal disruption.

P\&IDs include diagrammatic information on equipments (such as reactors, centrifuges, heat exchangers, pumps, etc), piping systems that transport materials and information, control instruments for measuring and regulating process variables, and actuating systems that adjust flows and pressures based on control signals.
Additionally, each P\&ID is also associated with one or many data sheets that detail out the material which is stored and processed in each pipe or equipment, along with its physical and chemical attributes.
The data sheets are generally standalone documents but can also be included in the diagram similar to Figure \ref{fig:PID_Example} (see left bottom corner).

\subsection{Data EXchange in the Process Industry (DEXPI)}
\label{bg:dexpi}

Engineers use different Computer Aided Engineering (CAE) software to view, edit and create P\&IDs.
However, each software tool has its own data handling framework and there exists a lack of interoperability between CAE (and other) systems \cite{wiedau-dexpi}. 
In executing such projects that encompass the planning, construction, and operation of process plants, companies encounter significant challenges in data exchange.
A primary contributor to these challenges is the lack of a universally adopted data exchange standard within the process industry, leading to incompatible systems and increased effort for data management \cite{DEXPI-specification}.
To enhance efficiency and digital interoperability, the \textit{Data EXchange in the Process Industry (DEXPI)} initiative was established.
The DEXPI initiative integrates concepts from various existing standards such as the  International Organization for Standardization (ISO),  International Electro technical Commission (IEC), BIM, Open Platform Communications Unified Architecture (OPC), Capital Facilities Information Handover Specification (CFIHOS) or NAMUR organization recommendations \cite{wiedau-dexpi}, and implements them based upon the ISO 15962 specification \cite{ISO15962}.
It offers a textual, machine-readable, and extendable representation of P\&IDs, facilitating interoperability.

The exchange format of the most recent DEXPI standard (DEXPI standard 1.3) is the Proteus Schema 4.1 \cite{Proteus4.1}, that ensures a mapping from DEXPI elements to XML patterns.
It provides an efficient way to encode P\&ID diagrams into XML files following the Proteus Schema.
A general DEXPI-compliant Proteus XML, referred to hereafter as the DEXPI XML, has the DEXPI Model as the root of the XML composition hierarchy. 
In the Proteus XML Schema the DEXPI Model is depicted as the \textit{Plant Model} XML Class and can be seen in Figure \ref{fig:results}. 
The DEXPI Model contains three important parts for a coherent and complete DEXPI XML representation of a P\&ID:
\begin{itemize}
    \item \emph{Conceptual Model}: This contains all the engineering information of the DEXPI model like elements, connections and their attributes, and is independent of the drawing.
    \item \emph{Drawing}: This class contains all the drawing information like the border, page dimensions or page color.
    \item \emph{Shape Catalogue}: This contains shape information of all the elements. It describes how shapes and symbols are created using lines and arcs.
\end{itemize}

\section{Architecture}
\label{arch}

The aim of the ACPID copilot is to create P\&IDs from Natural Language prompts. 
However, due to intricate design of P\&IDs, framing it as a Visual Generation task becomes difficult. 
Hence, we frame the problem as a Natural Language Generation task to create the DEXPI XML textual representation of the P\&ID from the prompts.
We then transform the textual representation into CAE based visual diagrams, rendered with Microsoft Visio, using deterministic rules.

We propose a novel workflow incorporating Plan and Execute Agents \cite{Wang2023PlanandSolvePI} and rule-based synthesis for the generation of P\&IDs. 
The agent-based system organizes and structures information in a format optimized for rule-based synthesis.
The output of the agent-based aggregator is then translated into the DEPXI Proteus XML representation through a deterministic, rule-based translation process.
The architecture of the ACPID copilot (Figure \ref{fig:arch}) can be divided into three main modules\textemdash the Textual Representation Creation, Visualization Generation, and Downstream tasks. The following subsections dive deeper into each of these three parts.

\subsection{Textual Representation Creation}
\label{method:concept}

The \textit{Textual Representation Creation} module translates natural language conversations describing P\&ID components into a DEXPI XML representation. 
This module's architecture is adapted from Microsoft's open-sourced Programming with Representation (PwR) framework \cite{ym2023pwr}, in which natural language prompts are converted to a Domain-Specific Language (DSL) and subsequently translated deterministically into code, enabling robust, interpretable, and efficient code generation. 
The intermediate DSL serves as a JSON equivalent of a Finite State Machine, outlining the workflow needed to integrate the element into the system. 
We modify the DSL generation and deterministic code generation components to meet the requirements for DEXPI generation.
The entire architecture is powered by a multi-step agentic workflow over a pre-trained LLM. 
The multi-step agentic workflow can be broken into the following main sub-modules:

\paragraph{Planning} The first step of the agentic workflow is creation of an execution plan from the user provided prompt. 
The planning step involves creation of steps that are to be performed by an LLM to add an element (equipment, instrument or actuating system) or a connection. 

\paragraph{Execution} Once the plan is generated, we iterate over the planned steps and execute each step using an LLM.
Each execution-step prompt contains the description of the step generated during the planning phase, and the utterance from the original user prompt that corresponds to the planned step. 
For each step, to provide context for existing elements and connections, we append outputs of all previously executed plan-steps with the prompt. 
The LLM which executes the plan can be different than the LLM which generates the plan. 

\paragraph{Validation and Pruning} When all the planned steps are executed, we deploy rule-based validation to validate the flow of the plan and execution steps, and prune unnecessary floating steps.
Currently, the validation only verifies the flow and transitions, but can be modified in the future to add P\&ID validation support for the generated content. 
The output of this step, consists of information derived from the prompt which is then expanded, formatted, and organized in a custom domain-specific language (DSL), which is essential for further processing.

\paragraph{Domain Specific Language (DSL)} The DSL output from the preceding sub-module acts as an intermediate representation before we convert it into the DEXPI XML. We propose that the conversion through an intermediate representation, like the DSL, that imposes fewer syntactic constraints compared to the final DEXPI XML, improves the quality of the translation. The DSL is specifically designed to encapsulate structured information derived from the input prompt, facilitating its organized and efficient rule-based conversion into DEXPI XML.

\paragraph{Translate to DEXPI} To generate the DEXPI XML representation of the P\&ID, we employ a deterministic, rule-based translation of the DSL. We employ predefined mapping, enabled by the encapsulation of information in the DSL, to improve consistency and reliability. Moreover, this determinism also helps improve the transparency and adaptability of the generation process.
 
\subsection{Visual Diagram Generator}
\label{method:visualization}
The preceding \textit{Textual Representation Creation Module} generates only the DEXPI Conceptual Model of the P\&ID, without including any visual elements. 
This textual output is subsequently utilized to construct the visual representation of the P\&ID. 
The \textit{Visual Diagram Generator} module then incorporates this visual information, enriching the DEXPI XML with data for the Shape Catalogue and the Diagram class.

The Conceptual Model generated by the Textual Representation Creation module is parsed and converted into a Microsoft Visio Drawing via the Microsoft Visio C\# API, serving as an initial draft for human validation and refinement. 
The \textbf{Human-in-the-Loop} (HITL) postulation in the workflow provides engineers with the ability to verify the accuracy of elements and connections within the generated P\&ID.
Additionally, HITL enables engineers to manually adjust the diagram according to project-specific preferences, personal preferences, and prior design standards.
Ultimately, HITL promotes robustness and responsible AI practices throughout the generation phase.

Once the engineer is satisfied with the Drawing, we extract the position and shape information and insert it into the existing DEXPI XML output from the first module.
We complete the DEXPI XML by adding position information to necessary elements and the shape information to the Drawing and Shape Catalogue classes in the DEXPI XML.

\subsection{Downstream Tasks and the Iterative Workflow}
\label{method:downstream}

The ACPID copilot is designed for subsystem-level generation, allowing engineers to iteratively loop through the entire workflow for adding more subsystems while efficiently reusing already generated subsystems. 
This approach emulates the current manual process, where engineers build the system step-by-step, focusing on one subsystem at a time. 
Subsystem-level generation also enables engineers to validate and refine their prompts more effectively, while optimizing the ACPID copilot’s performance by maintaining manageable prompt lengths \cite{levy-etal-promptLen}.
This capability for subsystem-level generation is what enables users to modify existing P\&IDs.

If no additional subsystems remain, our workflow enables engineers to perform various downstream tasks, including Question Answering, Root Cause Analysis (RCA), and summarization. 
This is due to the outputs and digital by-products generated throughout the workflow. 
The output of the Visualization Generation Module includes a Visual P\&ID, in the Visio Drawing format (\textit{.vsdx}), and a complete DEXPI XML including the Conceptual Model, Shape Catalogue and Drawing information.
During the Textual Representation Creation, we also obtain a Natural Language Description of the entire P\&ID, by parsing the DEXPI XML.
This description is generated for enabling multi-turn conversations during generation, and hence can be used optimally for downstream tasks as well.
As a result, due to the interoperability offered by DEXPI standards, and a spectrum of digital products created as outputs or by-products of the workflow, we enable multiple downstream tasks for the engineers.

\begin{figure*}[htbp]
    \centering
    \includegraphics[width=\linewidth]{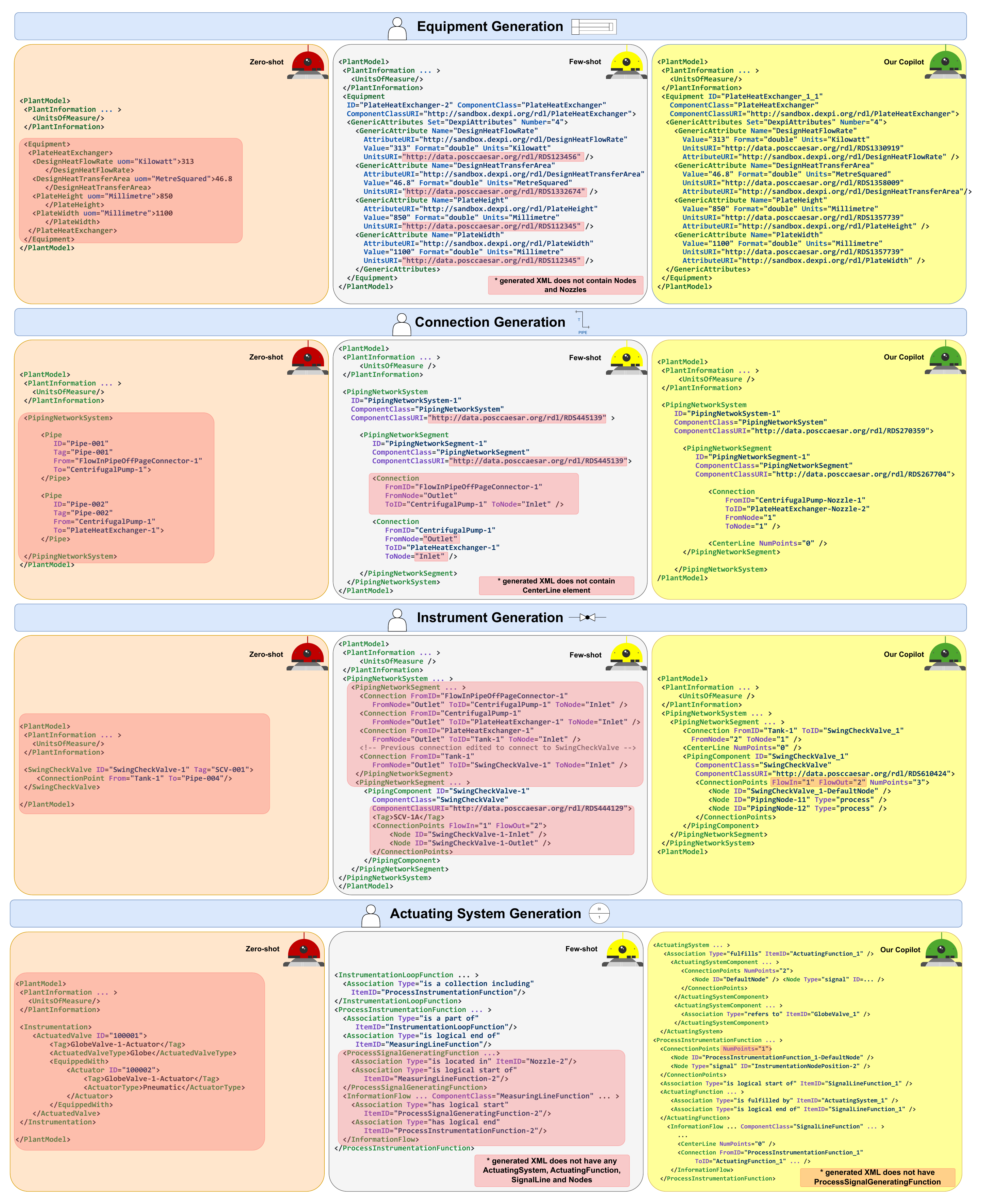}
    \caption{Illustrative outputs from zero-shot generation, few-shot generation, and our proposed Copilot. Errors are highlighted in red for clarity. Ellipsis indicate correct but lengthy content. The ACPID Copilot exhibits observable improvement in reliability, producing near-ground-truth examples of DEXPI XML. }
    \label{fig:results}
\end{figure*}

\section{Evaluations}
\label{eval}

\paragraph{Dataset:} To evaluate the performance of The ACPID copilot, we create an evaluation test-bench from the DEXPI examples \cite{dexpi-1.3-examples} provided by the DEXPI consortium. 
We create tests to evaluate the soundness of the generation and the completeness of the DEXPI Proteus XML creation.

\paragraph{Performance Metrics:} Aligned with its mathematical definition, we define the \textit{soundness} of the generation process as the property that ensures every element referenced in the prompt is included in the response in a structured and coherent manner.
Similarly, we define \textit{completeness} of the generation process, as the existence of all fields required for a DEXPI XML to be considered syntactically complete and to support interoperability.

\paragraph{Baseline:} We compare the results with zero-shot and few-shot performances of GPT-4-Turbo model which also acts as the base model for our copilot.
We create few-shot prompts by appending actual DEXPI XML outputs provided from the DEXPI Examples for elements and connections.
Care is taken to avoid using same examples during few-shot generation.
Figure \ref{fig:results} presents examples of outputs generated by The ACPID copilot, alongside those produced by the zero-shot and few-shot approaches for comparison.

\subsection{Soundness}
Mathematically, if $Elements$ is the set of all equipments, instruments and actuating systems, $Connections$ is a set describing connections between two elements and $Attributes$ is the set of all attributes for any connection or element, for an element $e \in Elements$, a connection $c \in Connections$ and an attribute $a \in Attributes$, we evaluate soundness as the proportion of inference-instances in which $e\ or\ c\ or\ a$  appears in both the prompt and the generated DEXPI XML. 
We report $Soundness$, as percentage responses that contain the element $e$ or the connection $c$ or the attribute $a$, whenever the corresponding prompt mentions $e\ or\ c\ or\ a$.

Our evaluation is based only on the DEXPI XML, as the visual diagram is a deterministic derivation of the XML in our workflow.
We extract about \textit{132} artifacts generated by elements, connections and attributes with a wide coverage over different scenarios, to evaluate soundness while generation.
We assess the soundness of the approach, irrespective of the completeness and correctness of the generated DEXPI XML.
Therefore, we also include incorrectly generated DEXPI XML samples and focus solely on evaluating whether the response incorporates the elements from the prompt in any structured manner.
We chose this evaluation approach based on the observation that both zero-shot and few-shot approaches fail to produce credible DEXPI XML. 

\begin{table}[h!]
\centering
\begin{tabular}{l|c}
\textbf{Method} & \textbf{Soundness} \\ \hline
Zero-Shot & 58.33\% \\ 
Few-Shot & 65.90\% \\ 
\textbf{ACPID Copilot} & \textbf{96.96}\% \\ 
\end{tabular}
\caption{Comparing soundness during generation, between Zero-shot, Few-shot generation from GPT-4-Turbo and the ACPID copilot.}
\label{tb:soundness}
\end{table}

The results can be seen in the Table \ref{tb:soundness}, where the ACPID copilot clearly outperforms the few-shot and zero-shot variations. 

\subsection{Completeness}
We identify sections of the DEXPI XML necessary to visualize and enable interoperability of the DEXPI XML across platforms. 
For equipments and instruments, we check the correctness of the XML Class (\textit{Element}), and its Attributes.
We then check the existence of the ID, the Component Name, the Component Class (and correctness of \textit{ComponentClassURL}), the Nozzles and Nodes, and the Position and Scale sub-elements.
For Connections, we check the correctness of the XML Class (\textit{PipingNetworkSystem, PipingNetworkSegment}), and the Connection element that specifies the source and destination element.
We then check the existence of CenterLine (element required for depiction of lines in the visual DEXPI XML).
Finally, for the actuating systems, we again evaluate the correctness of the XML Class (\textit{ActuatingSystem}) and check the existence of associated \textit{InstrumentationLoopFunction, ProcessInstrumentationFunction} with \textit{Actuating Function}, and \textit{InformationFlows} while verifying the correctness of necessary \textit{Associations} elements.  
We test the completeness by evaluating on \textit{555} sections extracted from a DEXPI Example provided by the DEXPI Consortium. 

\begin{table}[h!]
\centering
\begin{tabular}{l|c}
\textbf{Method} & \textbf{Completeness} \\ \hline
Zero-Shot & 0\% \\ 
Few-Shot & 68.28\%  \\ 
\textbf{ACPID copilot} & \textbf{92.97}\% \\ 
\end{tabular}
\caption{Comparing soundness during generation, between Zero-shot, Few-shot generation from GPT-4-Turbo and the ACPID copilot.}
\label{tb:completeness}
\end{table}

The results can be seen in the Table \ref{tb:completeness}, where our copilot outperforms the zero-shot and few-shot variations by a large margin. 

We observe that the vanilla zero-shot generation is entirely unable to create a complete and syntactically valid DEXPI XML even though GPT-4-Turbo has knowledge about DEXPI XML.
In contrast, we observe that the ACPID copilot outperforms both vanilla zero-shot and few-shot generation by a significant margin over the small test-bench.
We see that few-shot is able to match performance on simpler tasks like addition of Equipments and Attributes, but fails to extend a similar performance on more complicated tasks like addition of Connections, Instruments and Actuating Systems.
We can see some places where our copilot makes errors, but the source of such errors can be identified and corrected due to increased provenance of the entire generation process.

The high scores, of the ACPID copilot, in soundness and completeness can be attributed to a more rigid runtime framework because of the rule-based determinism.
Additionally, our experiments highlight another advantage: leveraging the Plan and Execute Agents enables our copilot to handle complex prompts—such as incorporating multiple elements and attributes or resolving ambiguous element references—where alternative methods prove inadequate.
Finally, our copilot's agentic ability to directly edit the XML eliminates the need to provide the entire current XML as context, as is required in zero-shot and few-shot generation approaches for XML completion, hence saving tokens to partially offset additional tokens required during Planning and Execution. 
However, there are some limitations which we discuss in the following section.

\section{Discussion and Future Work}
\label{future}

Even though the ACPID copilot outperforms the zero-shot and few-shot competitors, due to the rigidity of the rules, the prompts need to be constructed carefully for the entire workflow to run end-to-end.
However, this task can be automated in the future using Prompt Automation as depicted in the Figure \ref{fig:arch}.
Additionally, ensuring high accuracy alongside completeness and soundness during generation comes with the trade-off of increased inference time, in comparison to the other methods.
However, we estimate that our copilot results in a net reduction in the average time required to create a diagram in comparison to manual generation, by improving efficiency and productivity.

Also, a limitation of our study is the evaluation conducted on a small test bench, primarily due to the scarcity of standardized DEXPI XML data from real-world industrial plants. 
This limitation arises from the proprietary nature of majority of such data, which restricts its availability for research and validation purposes.

Finally, further enhancements could be made to our Copilot to make it more optimized with respect to the inference time and number of generated tokens.
Work can also be done to expand the current capabilities of the copilot to support other engineering diagrams, to improve the natural language processing aspects for better understanding of complex descriptions, and to integrate additional feedback mechanisms for continuous learning. 

\section{Conclusion}

This acts as a preliminary work in completely automating P\&ID generation.
In this work, we have presented a novel approach to generate P\&IDs directly from natural language descriptions, addressing a significant gap in the existing literature. 
Our proposed copilot leverages a multi-step agentic workflow to facilitate subsystem-level generation, enhancing the efficiency and accuracy of P\&ID creation.
To the best of our knowledge, ours is the first Generative AI based Designer Agent \cite{Batres-agent-pid} for P\&ID generation.
By incorporating Generative AI, we aim to alleviate the tedious and error-prone nature of manual diagram generation, ultimately streamlining the design process in the chemical and process industries.

Looking ahead, our research opens several avenues for future exploration. Our research also opens new avenues to create structured diagrams like architectural blueprints and circuit-boards by adopting our novel workflow, especially in data-constrained environments where knowledge from pretrained models can be exploited.
As Generative AI continues to evolve, we anticipate that our work will contribute to a more efficient and intelligent design process ultimately benefiting all stakeholders involved.

\bibliography{aaaiWorkshop}

\end{document}